\documentclass[runningheads]{llncs}
\usepackage[T1]{fontenc}
\usepackage{multirow} 
\usepackage{pifont} 
\usepackage{graphicx}
\usepackage{xcolor}
\usepackage{booktabs}
\usepackage{makecell}
\usepackage{subcaption}
\usepackage{caption}

\definecolor{trolleygrey}{rgb}{0.5, 0.5, 0.5}
\definecolor{trolleygrey}{RGB}{128, 128, 128}
\definecolor{softgray}{RGB}{200, 200, 200}

\newcommand{\seg}[1]{\noindent \textbf{#1}~}

\usepackage{amsmath,amsfonts,bm}

\def\eqref#1{equation~\ref{#1}}

\def\1{\bm{1}}

\def\vx{{\bm{x}}}

\def\mC{{\bm{C}}}

\def\mI{{\bm{I}}}

\def\mQ{{\bm{Q}}}

\def\mS{{\bm{S}}}

\def\mW{{\bm{W}}}

\DeclareMathAlphabet{\mathsfit}{\encodingdefault}{\sfdefault}{m}{sl}
\SetMathAlphabet{\mathsfit}{bold}{\encodingdefault}{\sfdefault}{bx}{n}

\usepackage{hyperref}
\usepackage{graphicx,verbatim}
\begin{document}
\title{BREIT: A Framework for \underline{B}rain Stroke \underline{R}econstruction using  Multi-Frequency 3D \underline{EIT}}
\titlerunning{BREIT: A Framework for Brain Stroke Reconstruction using 3D MF-EIT}
\author{Djahid Abdelmoumene\inst{1,2}\and Ishak Ayad\inst{3}\and Ma\"{\i} K. Nguyen\inst{1}\and \\ Christian Daveau\inst{2}
}
\authorrunning{D. Abdelmoumene et al.}
\institute{ETIS (UMR 8051), CY Cergy Paris University, ENSEA, CNRS, Cergy France \and
AGM (UMR 8088), CY Cergy Paris University, CNRS, Cergy, France \and
Paris-Saclay University, CentraleSup\'elec, Inria, CVN, Gif-sur-Yvette, France \\
\email{djahid.abdelmoumene@cyu.fr}
}

\maketitle              

\begin{abstract}
Multi-Frequency Electrical Impedance Tomography (MF-EIT) is a non-invasive, low-cost modality that reconstructs electrical property distributions from boundary voltages. For stroke imaging, progress in 3D deep-learning reconstruction is limited by the lack of large-scale datasets with paired ground-truth (GT) volumes and by non-standardized pipelines for data generation, simulation, and evaluation. We introduce \textit{BREIT}, a modular framework for 3D MF-EIT stroke reconstruction providing: (i) a neuroimaging-to-EIT pipeline that converts CT/MRI into frequency-dependent GT admittivity volumes; (ii) a self-contained Python 3D Complete Electrode Model (CEM) forward solver for simulating MF-EIT voltages; and (iii) a 3D D-bar implementation supporting non-uniform electrode layouts. Building on BREIT, we propose \textit{dFNO-bar}, which integrates Fourier Neural Operators into D-bar by learning a mapping from scattering data $t(\xi)$ to conductivity $\sigma(x){=}\Re\{\gamma\}$. We evaluate dFNO-bar against D-bar, Deep D-bar, and Gauss--Newton reconstructions on UCLH-matched synthetic data, and observe higher brain SSIM with comparable CC across noise settings.
\keywords{Multi-frequency EIT \and Stroke reconstruction \and Fourier neural operator \and Deep Learning}
\end{abstract}

\section{Introduction and Related Works}
Rapid stroke assessment is critical because ischemic and hemorrhagic strokes require different treatments and have short therapeutic windows \cite{saver2006time,goren2018data}. While CT/MRI are the clinical gold standard, they are costly, infrastructure-dependent, and often unavailable for pre-hospital, bedside, or resource-limited use \cite{yuen2022portablemri}. In addition, stroke pathology can evolve over minutes to hours, whereas conventional imaging provides only intermittent snapshots \cite{schwamm1998serial}.

Multi-frequency Electrical Impedance Tomography (MF-EIT) is a promising complementary modality: it is non-invasive, inexpensive, portable, and enables continuous monitoring via scalp electrodes \cite{adler2017eit}. MF-EIT exploits frequency-dependent tissue admittivity to enhance contrast between brain tissues \cite{zhou2015tv}.

EIT stroke studies using ML/DL have largely targeted stroke-type identification from measurements or from compact, engineered representations. Approaches include learning directly from boundary-measurement operators versus training on noise-robust, geometrically motivated features, typically using simulated data \cite{Agnelli2020}. Classification has also been studied on 3D head models with generated synthetic datasets and systematic perturbations to assess robustness \cite{candiani2022nn}. More recent work further emphasizes realistic numerical head modeling and validation protocols for applied stroke differentiation studies \cite{culpepper2023ml}. Learning-based image reconstruction has also been explored, but remains dominated by 2D simplified workflows. Generative or convolutional models have been trained to map measurements to conductivity images using numerically generated 2D datasets \cite{ivanenko2024gan}. Other reconstruction networks, such as cascaded CNNs and attention/residual U-Net variants, have been proposed for stroke EIT imaging to improve artifact suppression and reconstruction quality under simulated conditions \cite{Liu2023,Liu2024}. Clinically grounded benchmarking is supported by the public UCLH dataset release pairing MF-EIT with CT/MRI \cite{goren2018data}, where \cite{mcdermott2020mf} used it for MF-EIT reconstruction, using a symmetry-difference technique combined with classical classifiers like SVMs to distinguish and reconstruct different cases, but overall the literature still varies substantially in modeling assumptions and simulation details, complicating reproducible comparison across pipelines.

Beyond data realism, reproducibility is hindered by inconsistent choices in phantom design, admittivity tables, electrode models, and forward modeling settings, making cross-paper comparisons difficult. Many pipelines also stop at forward simulation and do not provide integrated tooling for preprocessing, frequency mapping, training, and systematic evaluation. In contrast, BREIT unifies radiology-based admittivity generation, realistic MF-EIT simulation, and DL benchmarking under consistent configurations, enabling reproducible comparisons across stroke types and datasets.
To address these gaps, we introduce BREIT, a framework that generates anatomically grounded, 3D, frequency-dependent admittivity volumes from CT / MRI and simulates MF-EIT measurements using the Finite Element Method (FEM) with the 3D Complete Electrode Model (CEM) on realistic head meshes. We further propose \textit{dFNO-bar}, which integrates Fourier Neural Operators \cite{fno} into a D-bar reconstruction pipeline. Our main contributions are:
\begin{enumerate}
    \item \textbf{Multimodal data generation:} CT/MRI-based tissue/lesion mapping to 3D frequency-dependent admittivity volumes, paired with simulated MF-EIT voltages.
    \item \textbf{3D CEM + D-bar tooling:} a Python implementation of 3D D-bar components (including $t^{\mathrm{exp}}$ and $t^{0}$ approximations) for non-uniform electrode layouts \cite{Hamilton2021EIT3DtexpCalderon,Delbary2012EIT3DScatteringTransforms}.
    \item \textbf{dFNO-bar:} a 3D MF-EIT reconstruction model that couples FNO learning with D-bar structure.
    \item \textbf{Modular codebase:} reproducible generation, training, and evaluation across stroke types and datasets.
\end{enumerate}

\begin{figure}[!t]
    \centering
    \includegraphics[width=\linewidth]{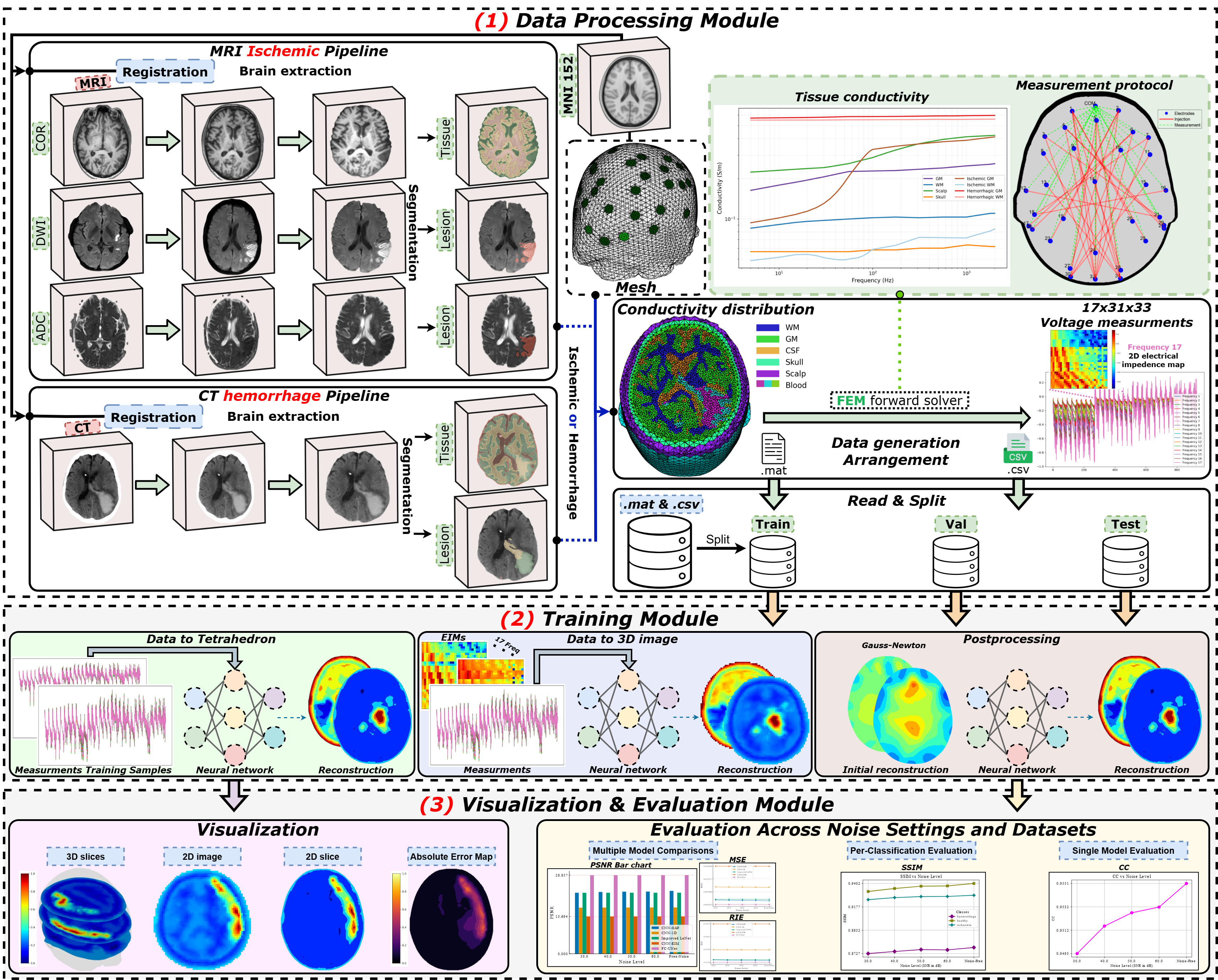}
    \caption{\textbf{The general structure} of the modular-based codebase of \textit{BREIT}.}
    \label{fig:pipeline}
\end{figure}

\section{Methodology}
\label{sec:method}

\subsection{BREIT Pipeline}
\label{sec:breit}
\subsubsection{Data Generation.}
\label{sec:data}
We base data generation on the UCLH MF-EIT stroke release~\cite{goren2018data}, a public cohort with clinically measured MF-EIT paired with CT/MRI. To enable subject-independent modeling while preserving UCLH measurement conditions (electrode geometry and frequency protocol), all imaging is registered to MNI152, converted to tissue/lesion segmentations in this common space, and projected to head meshes used for simulation and reconstruction. We use two data streams: (i) anatomically grounded admittivity maps derived from UCLH neuroimaging, and (ii) additional admittivity maps from external MRI/CT cohorts processed with the same pipeline and forward-simulated to augment training data.

\subsubsection{Ground-Truth Approximation Pipeline.}
For each subject, we form a 4-class segmentation (GM, WM, CSF, and lesion) and assign frequency-dependent complex admittivities using literature values \cite{Horesh2006} over 5--2000\,Hz (17 frequencies), yielding 17 admittivity volumes. All images are registered to MNI152 to enforce a common anatomical frame and to provide consistent head/skull/scalp geometry for meshing and electrode placement.

\seg{(a) Ischemic Stroke:}
We use DWI, ADC, and a high-resolution structural MRI. Volumes are rigidly aligned to MNI152 (FSL FLIRT \cite{FSL}), followed by brain extraction (FSL BET \cite{FSL}) and tissue segmentation of the structural MRI into GM/WM/CSF (FSL FAST \cite{FSL}). Ischemic lesions are segmented from the DWI/ADC pair using ADS \cite{ads}; the lesion mask is overlaid with the tissue map to form the final 4-label segmentation.

\seg{(b) Hemorrhagic Stroke:}
We process the clinical CT by registering to MNI152 (ANTs \cite{ANTS}), followed by brain extraction (CT\_Bet \cite{CTBet}) and tissue segmentation into GM/WM/CSF (CTSeg \cite{brudfors2020flexible}). Hemorrhage is segmented using DeepBleed \cite{Sharrock2021}; the hemorrhage mask is aligned to the tissue map and overlaid to create the 4-label segmentation (GM, WM, CSF, hemorrhage). The same frequency-dependent admittivity assignment is then applied.
\subsubsection{Data Augmentation.}
Because UCLH contains a limited number of measured subjects, we augment training by running the same neuroimaging-to-admittivity pipeline on external cohorts, ischemic MRI cases \cite{HernandezPetzsche2022}, public hemorrhage head CT cohorts \cite{bhsd,Hssayeni2020PhysioNetCTICH} and a healthy cohort \cite{ixi_dataset}. We then forward simulate MF-EIT voltages under the UCLH-matched frequency protocol and electrode geometry. All volumes are registered to MNI152 and rasterized onto a tetrahedral head mesh; we use a fine mesh for forward simulation ($\approx$1M tets) and a coarser mesh for reconstruction ($\approx$75k).

\subsubsection{Forward Problem Simulation.}
Given each subject-specific, frequency-dependent admittivity field $\gamma(\omega,\vx)=\sigma(\vx)+i\omega\epsilon(\vx)$ on the fine tetrahedral head mesh, we forward-simulate MF-EIT voltages using the 3D Complete Electrode Model. We replicate the UCLH electrode geometry and injection protocol across the 17 frequencies. In practice, the resulting electrode voltages are defined only up to an additive constant shift, so comparisons and learning objectives are insensitive to global offsets. We simulate with a fixed current amplitude since voltages scale linearly with injected current. Using a standard P1 tetrahedral FEM discretization of the CEM model we obtain the usual CEM block linear system~\cite{Vauhkonen1999ThreeDEITCEM}. We solve the augmented-real system with PyPardiso. We validated our Python forward solver by comparing simulated voltages against EIDORS~\cite{eidors} under matched meshes, electrodes, and protocols.

\subsubsection{Pipeline validation.} 
\begin{figure}[t]

\begin{minipage}{0.67\textwidth}
\centering

\begin{subfigure}{\textwidth}
\centering
\includegraphics[width=\linewidth,trim=8pt 0pt 0pt 0pt,clip]{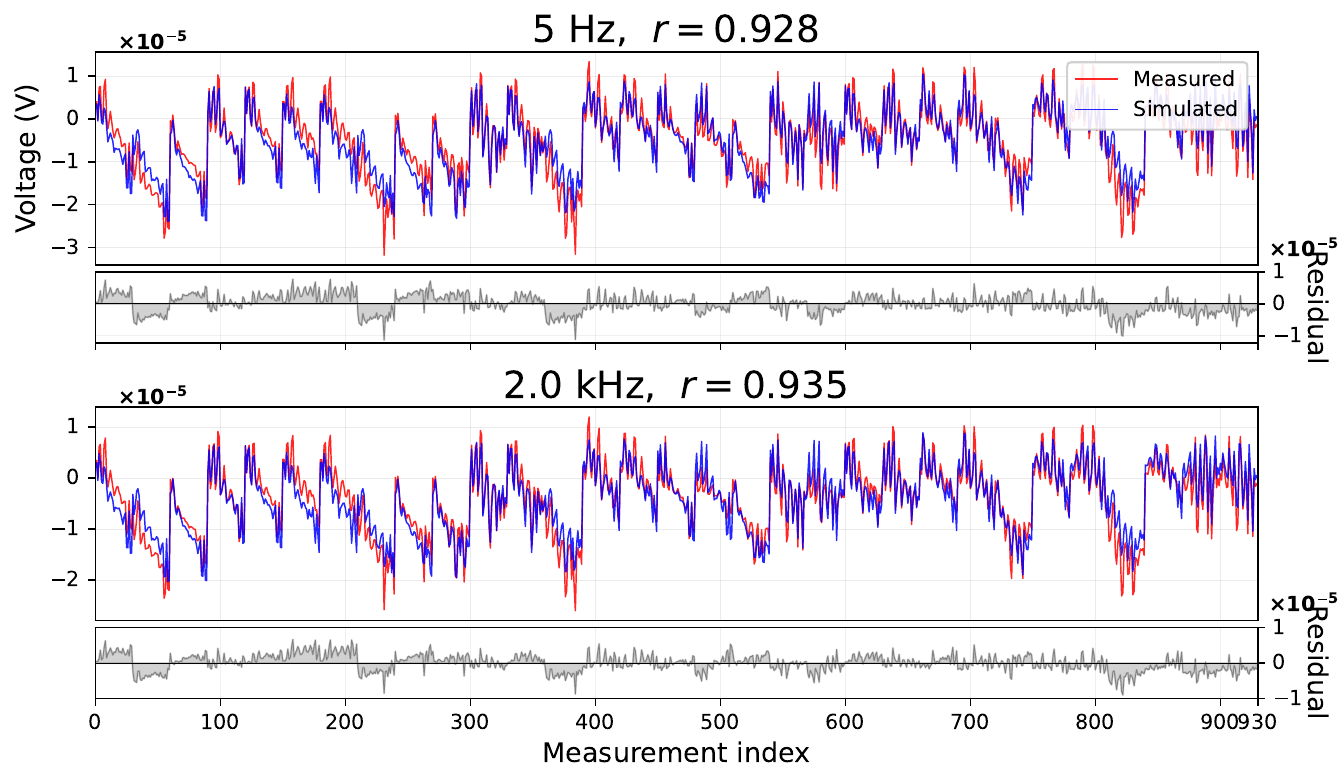}
\end{subfigure}

\caption{\textbf{Forward-solver voltage agreement with clinical MF-EIT}. 
Overlays for pat-01 at 5\,Hz and 2\,kHz after per-frequency affine alignment.}
\label{fig:overlay}

\end{minipage}\hfill
\begin{minipage}{0.30\textwidth}
\centering
{\fontsize{8}{9}\selectfont
\captionof{table}{\textbf{Per-patient mean $\bar r$ and NMSE} across 17 frequencies after per-frequency affine alignment.}

\begin{tabular}{lcc}
\toprule
Pat. & $\bar r$ & NMSE \\
\midrule
pat-01     & 0.928 & 0.103 \\
pat-03     & 0.895 & 0.144 \\
pat-04$_a$ & 0.909 & 0.140 \\
pat-04$_b$ & 0.894 & 0.170 \\
pat-05     & 0.926 & 0.106 \\
pat-06$_a$ & 0.895 & 0.161 \\
pat-06$_b$ & 0.745 & 0.428 \\
\dots      & \dots & \dots \\
\midrule
\textbf{Mean} & \textbf{0.817} & \textbf{0.290} \\
\bottomrule
\end{tabular}
\label{tab:overlay_metrics}
}
\end{minipage}

\end{figure}

We validate simulation realism by comparing forward-simulated voltages from our pipeline to the corresponding clinical MF-EIT measurements in UCLH for all subjects with matched imaging. In total, 12 patients (14 sessions)  were included, and agreement is summarized in Fig.~\ref{fig:overlay} and Table~\ref{tab:overlay_metrics}.

\begin{figure}[t]
    \centering
    \includegraphics[width=\linewidth]{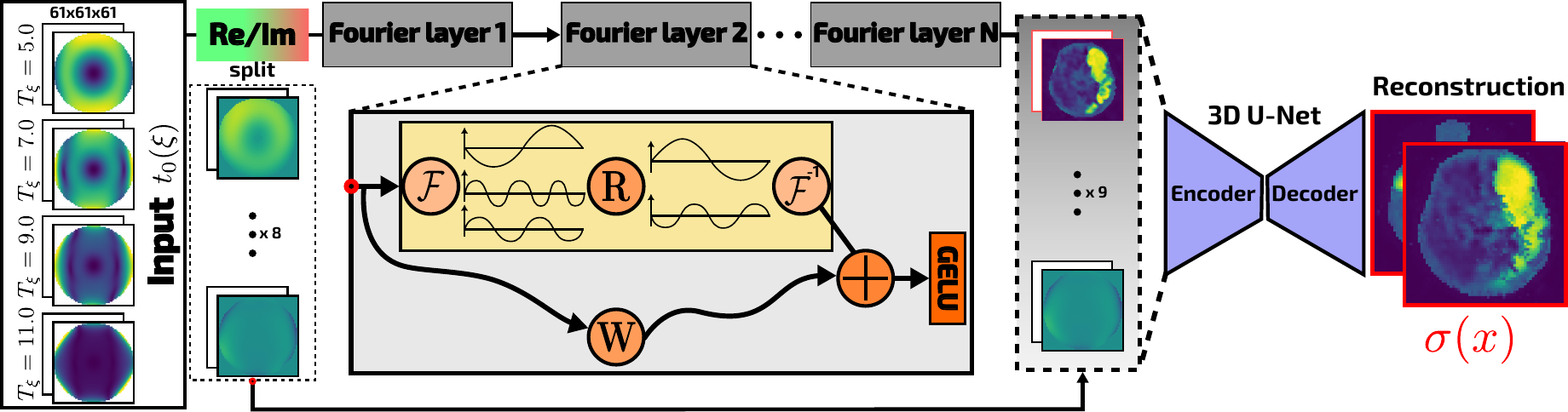}
    \caption{\textbf{Proposed model architecture.} A FNO maps the input in Fourier space (split real/imag) through stacked Fourier blocks, followed by a lightweight U-Net decoder to produce the spatial-domain output conductivity $\sigma(x)$.}
    \label{fig:model_arch}
\end{figure}

\subsubsection{Benchmarking.}
\label{sec:benchmarking}
\textit{BREIT} is an end-to-end, configurable benchmark stack spanning data loading, reconstruction/training, and visualization/evaluation (Fig.~\ref{fig:pipeline}). Experiments are driven by a shared configuration (e.g., resolution, optimizer, noise), with standardized multi-frequency loaders and mesh$\leftrightarrow$voxel conversion to ensure consistent comparisons across methods and datasets.

\subsection{The proposed dFNO-bar}

\subsubsection{3D D-bar with non-uniform electrodes.}
We implement the 3D electrode-data D-bar reconstruction in Python, following~\cite{Hamilton2021EIT3DtexpCalderon} for the numerical electrode-data pipeline and~\cite{Delbary2012EIT3DScatteringTransforms} for the $t_0$ scattering approximation. To better handle non-uniform electrode distributions, we replace the uniform boundary quadrature $4\pi/L$ where $L$ is the number of electrodes, with spherical Voronoi surface-area weights computed from the electrode center locations. Let $\vx_\ell\in S^2$, $\ell=1,\dots,L$, denote the electrode centers projected to the unit sphere, and let $w_\ell$ be the corresponding Voronoi areas, renormalized such that $\sum_{\ell=1}^L w_\ell = 4\pi$; We define $\mW=\mathrm{diag}(w_1,\dots,w_L)\in\mathbb{R}^{L\times L}$. These weights induce a consistent discrete boundary inner product, which we use in a weighted modified Gram--Schmidt step to orthonormalize the applied current patterns,
$\mC=\mQ\mS$ with $\mQ^\top\mW\mQ=\mI_m$, where $\mI_m$ is the $m\times m$ identity
matrix and $m$ is the number of independent current patterns.
We then follow the remaining steps of~\cite{Hamilton2021EIT3DtexpCalderon} and evaluate the $t_{exp}$ scattering approximation using the weighted quadrature:
\begin{equation}
\label{eq:texp_weighted}
t_{\exp}(\xi)
\;\approx\;
\big[e^{-i\vx\cdot(\xi+\zeta)}\big]^{\top}\,
\mW\,
\mQ\,dL\,(\mQ^{\top}\mW)\,
\big[e^{i\vx\cdot\zeta}\big],
\end{equation}
where $dL$ denotes the resulting DN-map difference in the $Q$-basis.
Scattering data are evaluated on a truncated Cartesian $\xi$-grid $\{\xi:\|\xi\|\le T_\xi\}$, and for each $\xi$ we choose $\zeta(\xi)$ as in~\cite{Hamilton2021EIT3DtexpCalderon}.
For the $t_0$ approximation~\cite{Delbary2012EIT3DScatteringTransforms}, we compute the boundary trace $\psi^0(\cdot,\zeta)$ by solving the discretized boundary integral equation $\big(\mI + \mS_0\,\mQ\,dL\,(\mQ^\top \mW)\big)\,\psi^0 \;{=}\;  e^{i \vx_\ell\cdot\zeta},$
where $S_0$ is the Laplace single-layer operator discretized on electrode centers, $(\mS_0)_{\ell j}\approx\frac{w_j}{4\pi\|x_\ell-\vx_j\|}$. We then evaluate $t_0$ by replacing $\big[e^{i\vx\cdot\zeta}\big]$ with $\psi^0$ in Eq.~\ref{eq:texp_weighted}. Finally, $q$ is obtained by inverse FFT of $t(\xi)$, and $\sigma_{\mathrm{rec}}$ is recovered as in \cite{Hamilton2021EIT3DtexpCalderon}.

\subsubsection{dFNO-bar architecture.}
We propose dFNOBar, a learned reconstruction method operating on 3D D-bar scattering data. We compute the complex scattering $t_0(\xi)$ at multiple truncation radii $T_\xi \in \{5.0, 7.0, 9.0, 11.0\}$, where smaller $T_\xi$ emphasizes low-frequency structure and larger $T_\xi$ retains higher-frequency detail. We stack the real and imaginary parts across truncations to form an 8-channel input volume. An FNO first predicts a coarse conductivity volume $\hat{\sigma}_{\mathrm{FNO}}(x)$ from this multi-$T_\xi$ scattering input. A lightweight 3D U-Net~\cite{ronneberger2015unet} then refines the prediction by concatenating it with the original input, giving 9 channels in total. This enables local refinement of the coarse output. The dFNO-bar architecture is shown in Fig.~\ref{fig:model_arch}.

\begin{figure}[t]
\centering
\setlength{\tabcolsep}{1.5pt} 
\renewcommand{\arraystretch}{0.9} 
\footnotesize
\begin{tabular}{cccccc}
  & One step GN\cite{noser} & D-bar\cite{nachman1988reconstructions} & Deep D-bar\cite{deepdbar} & \textbf{dFNO-bar} & Ground-truth \\
  \raisebox{1.\height}{\rotatebox{90}{Axial}} &
  \includegraphics[width=0.178\textwidth]{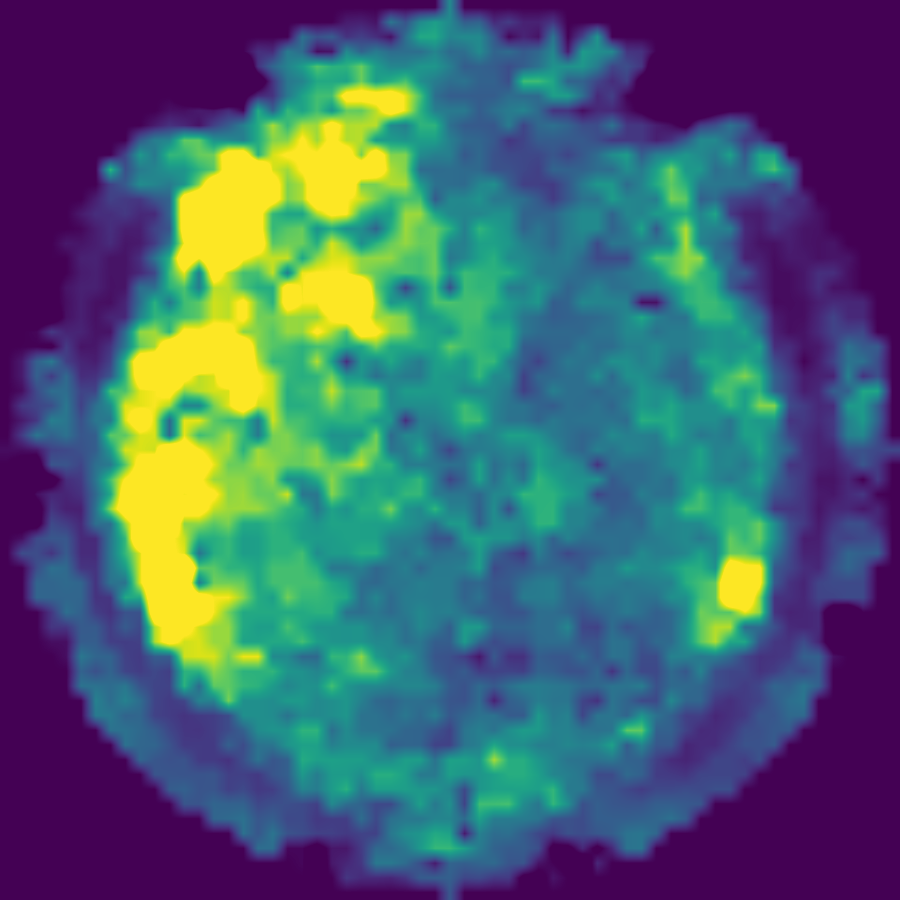} &
  \includegraphics[width=0.178\textwidth]{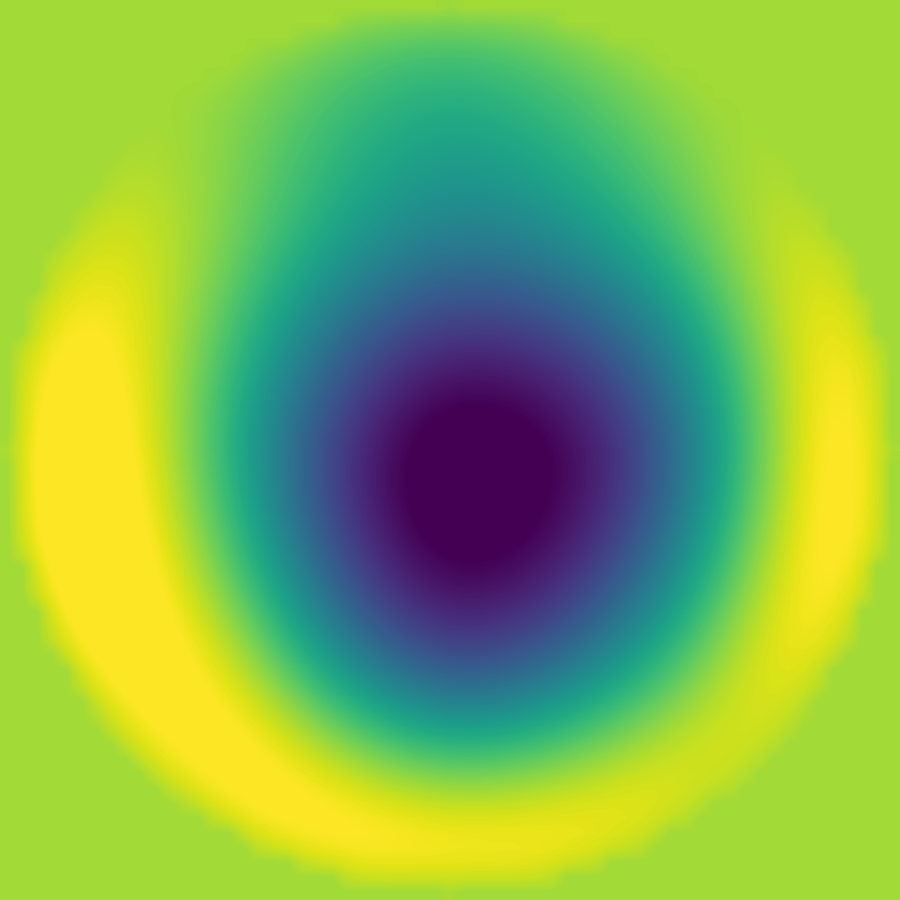} &
  \includegraphics[width=0.178\textwidth]{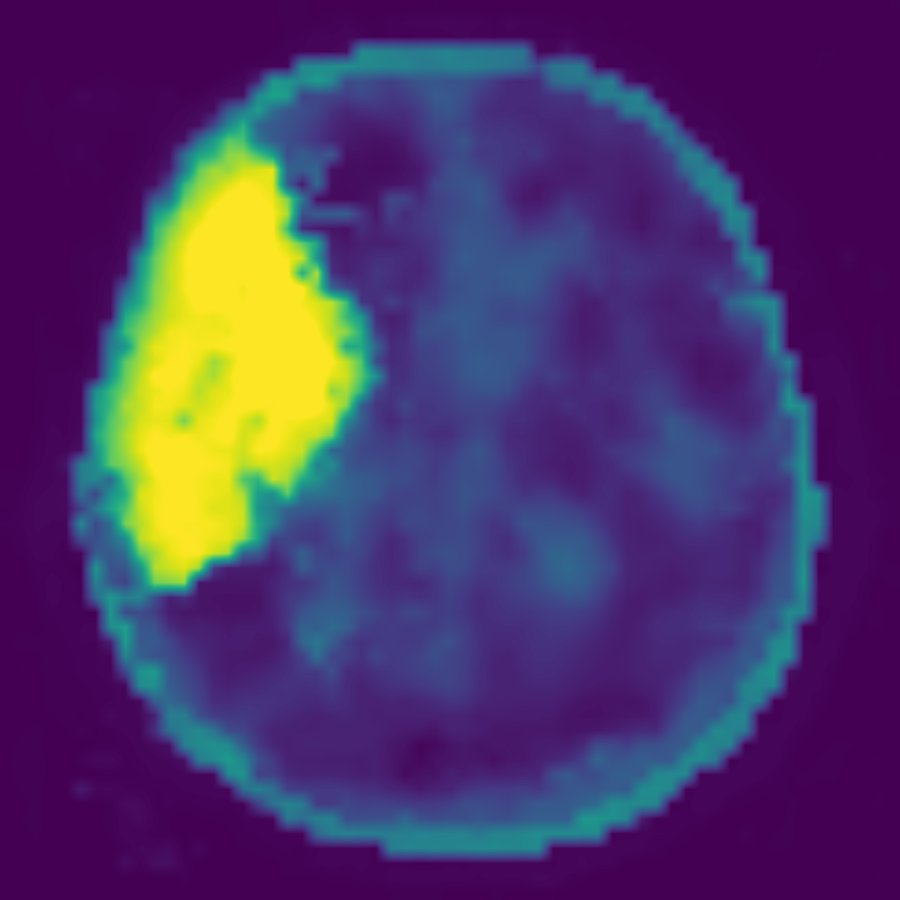} &
  \includegraphics[width=0.178\textwidth]{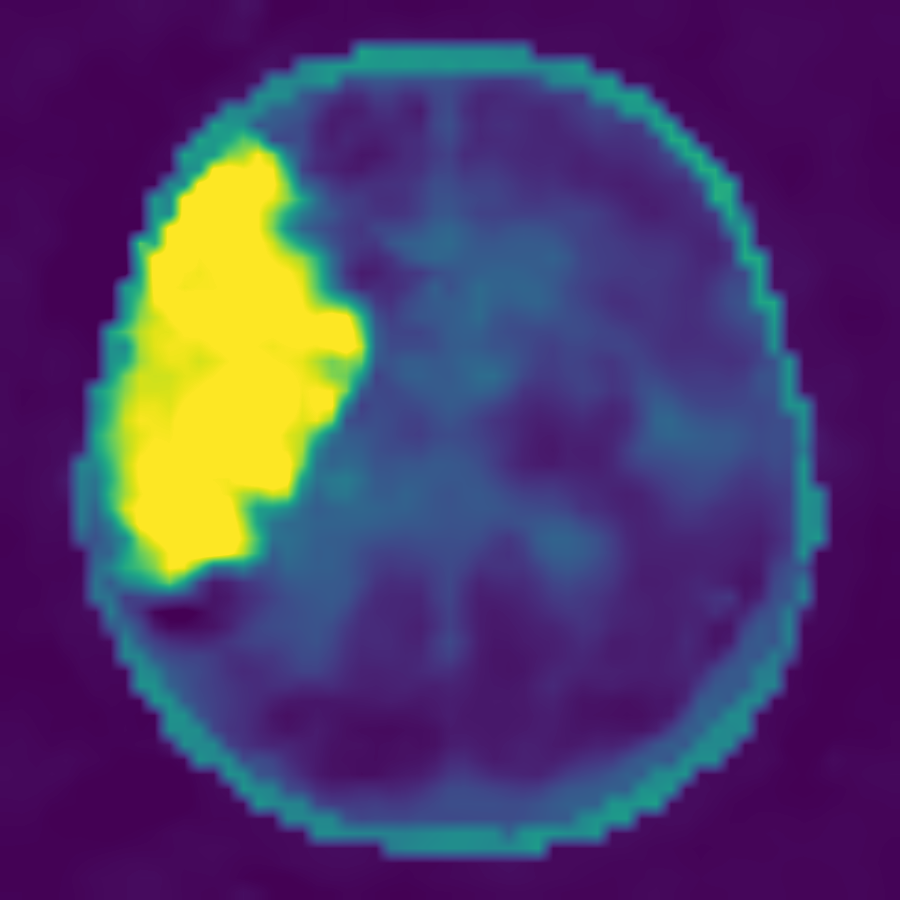} &
  \includegraphics[width=0.178\textwidth]{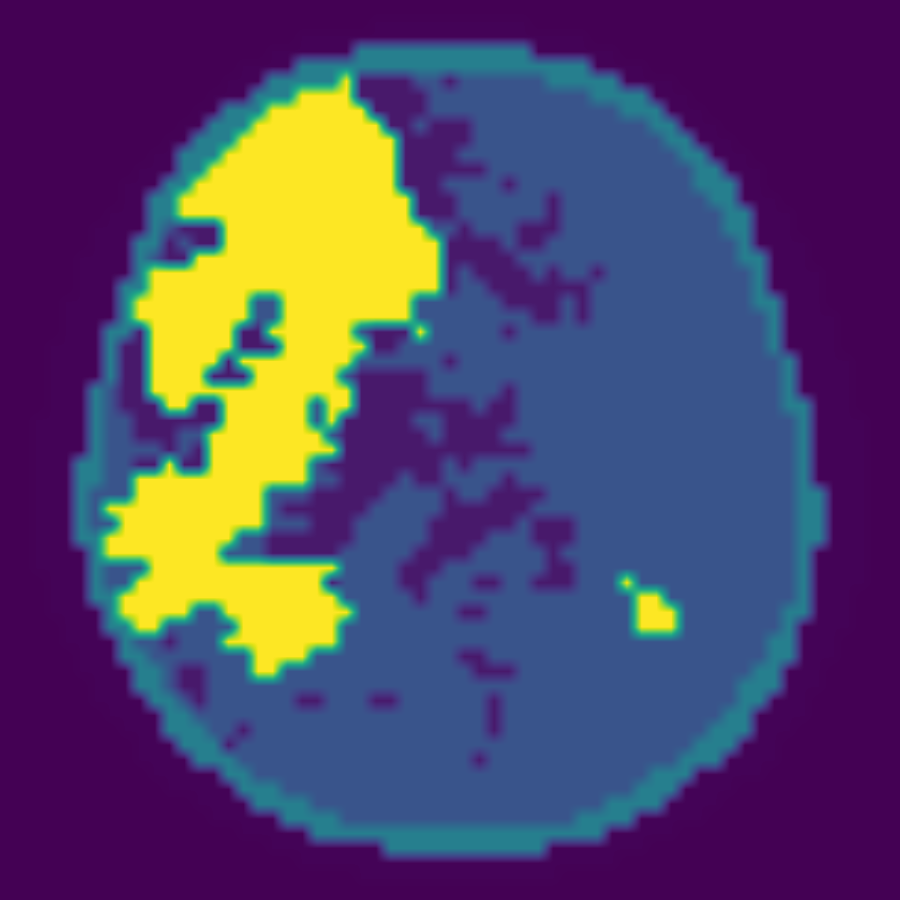} \\
  \raisebox{.5\height}{\rotatebox{90}{Sagittal}} &
  \includegraphics[width=0.178\textwidth]{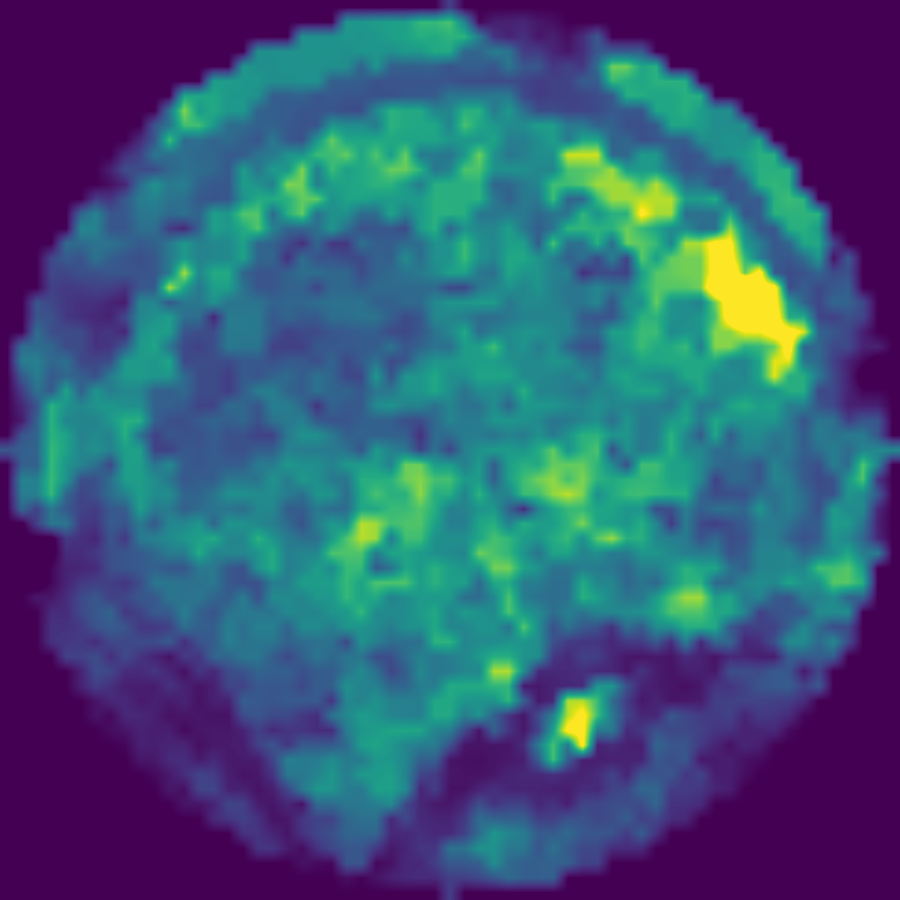} &
  \includegraphics[width=0.178\textwidth]{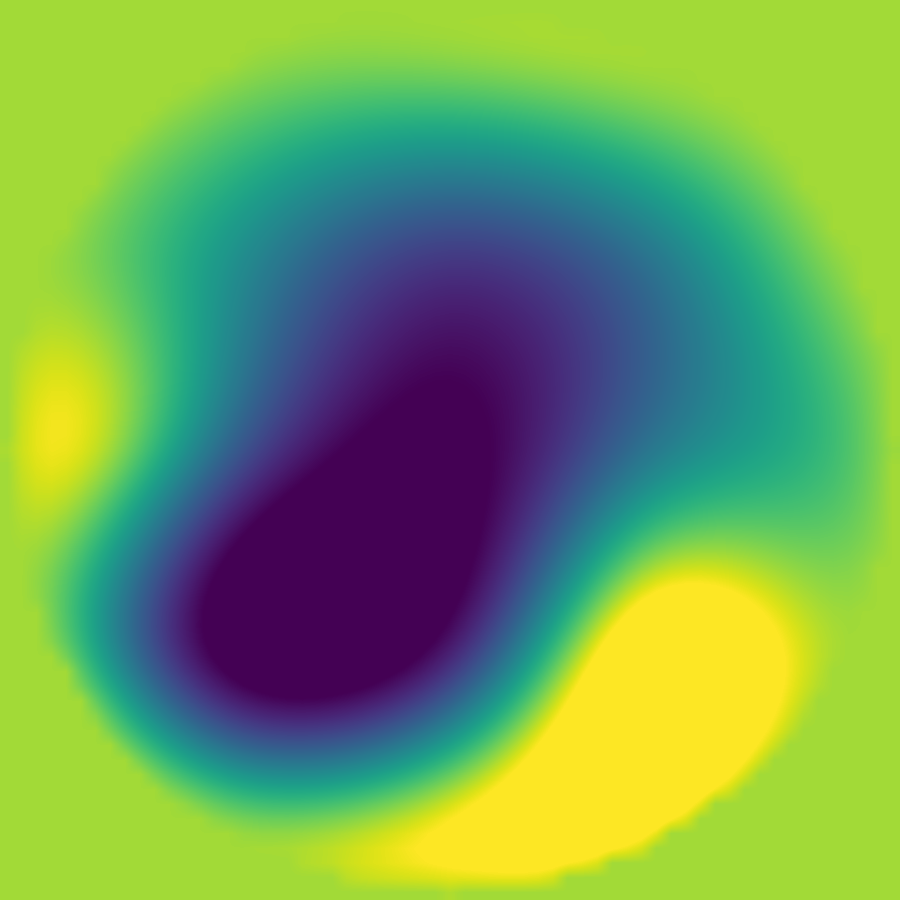} &
  \includegraphics[width=0.178\textwidth]{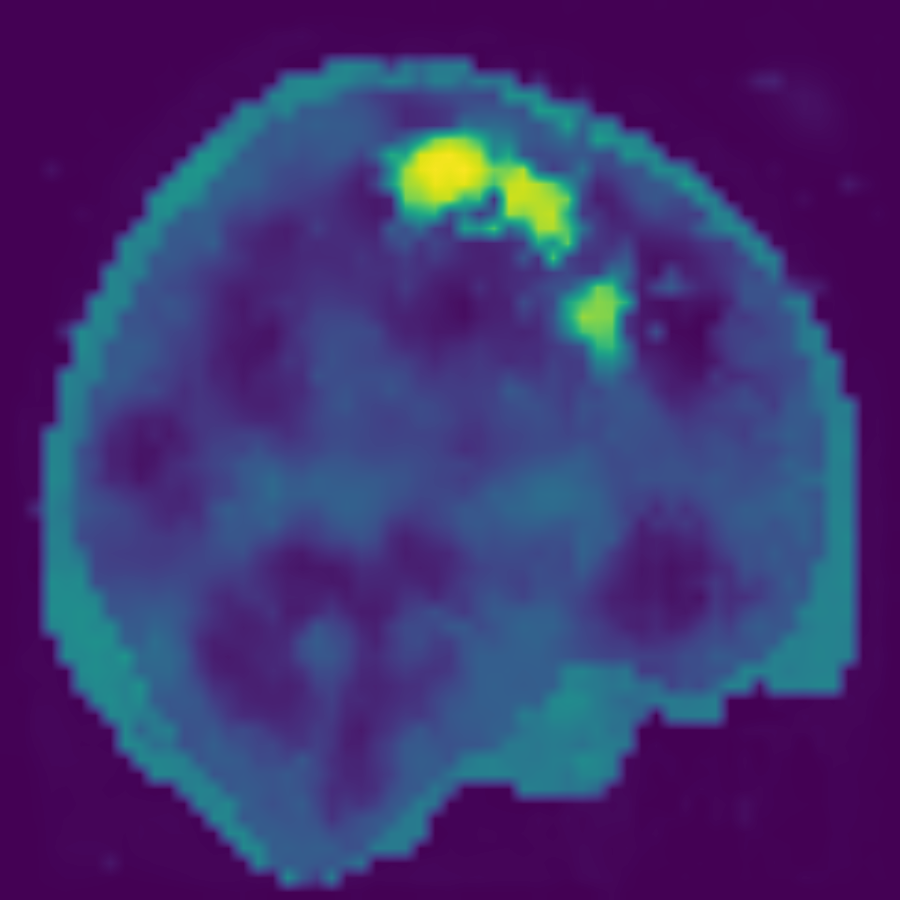} &
  \includegraphics[width=0.178\textwidth]{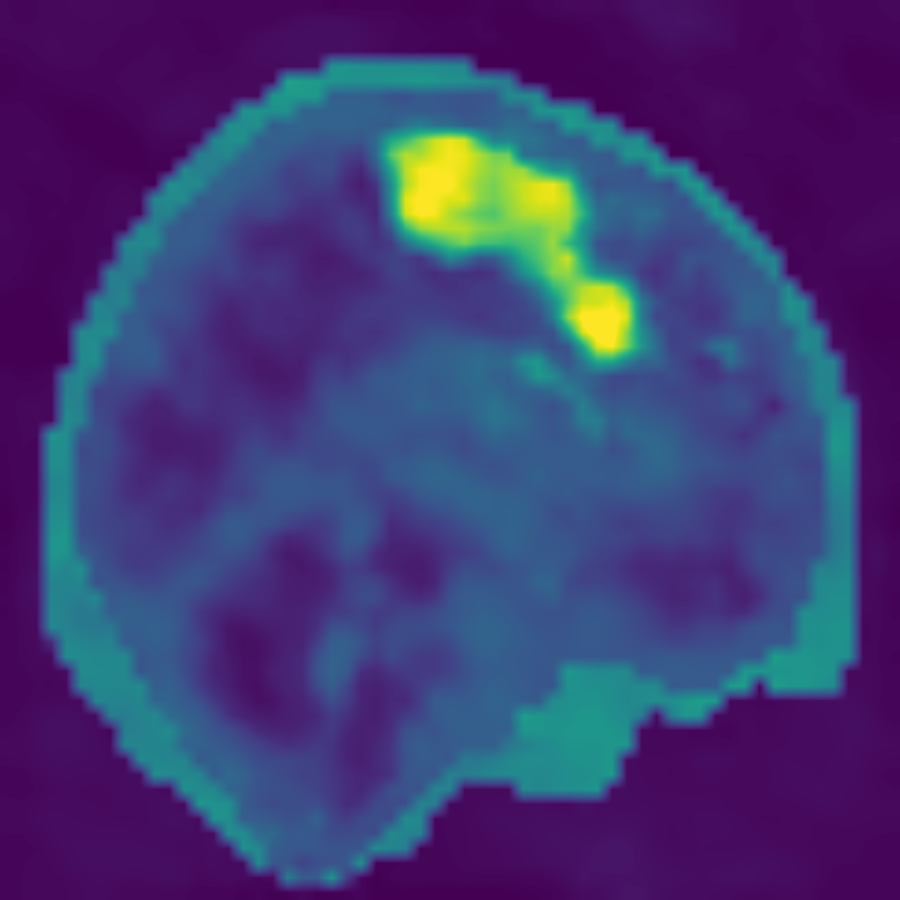} &
  \includegraphics[width=0.178\textwidth]{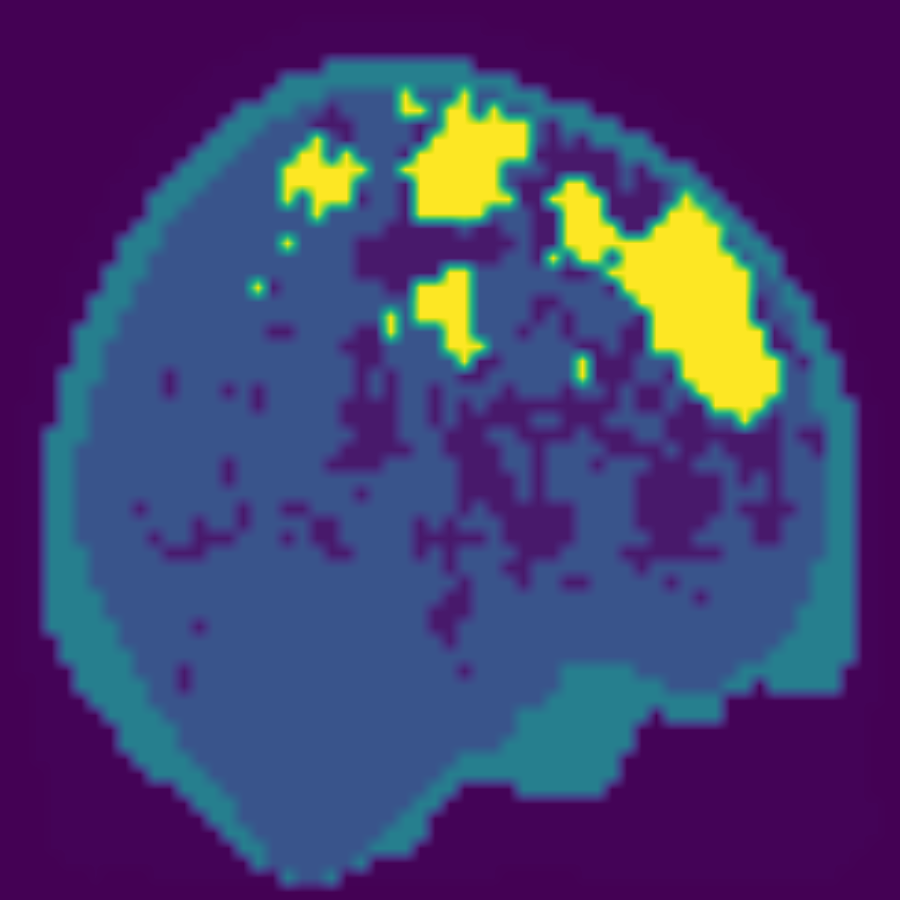} \\
  \raisebox{.5\height}{\rotatebox{90}{Coronal}} &
  \includegraphics[width=0.178\textwidth]{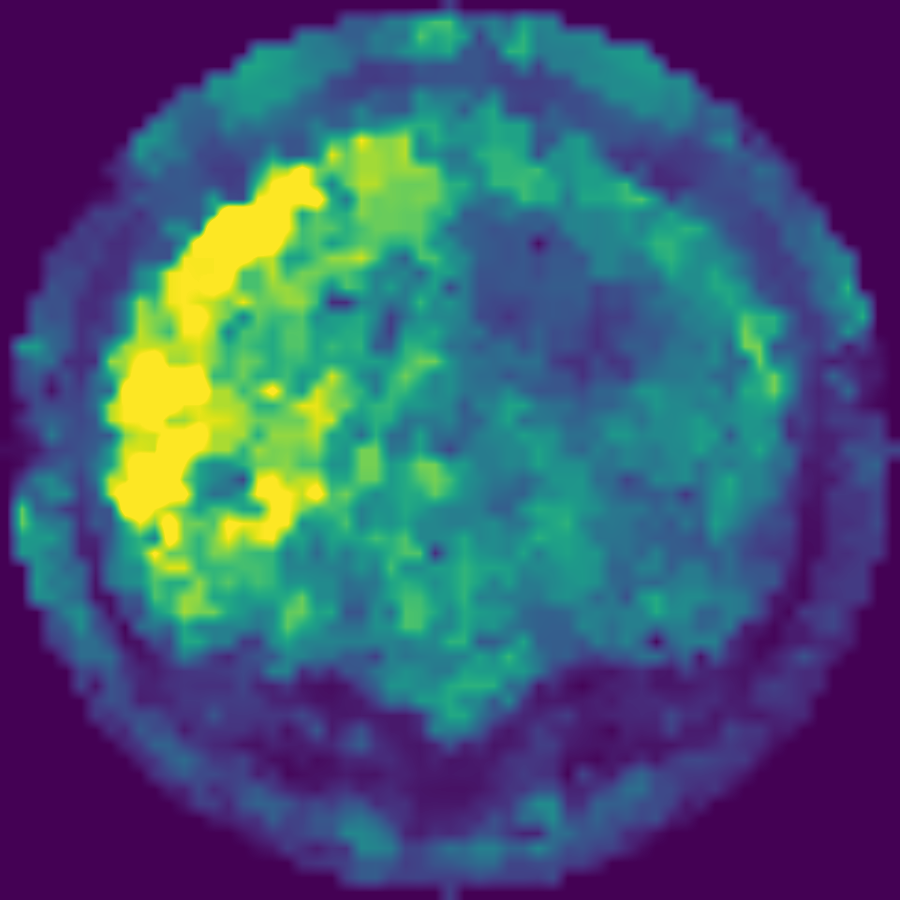} &
  \includegraphics[width=0.178\textwidth]{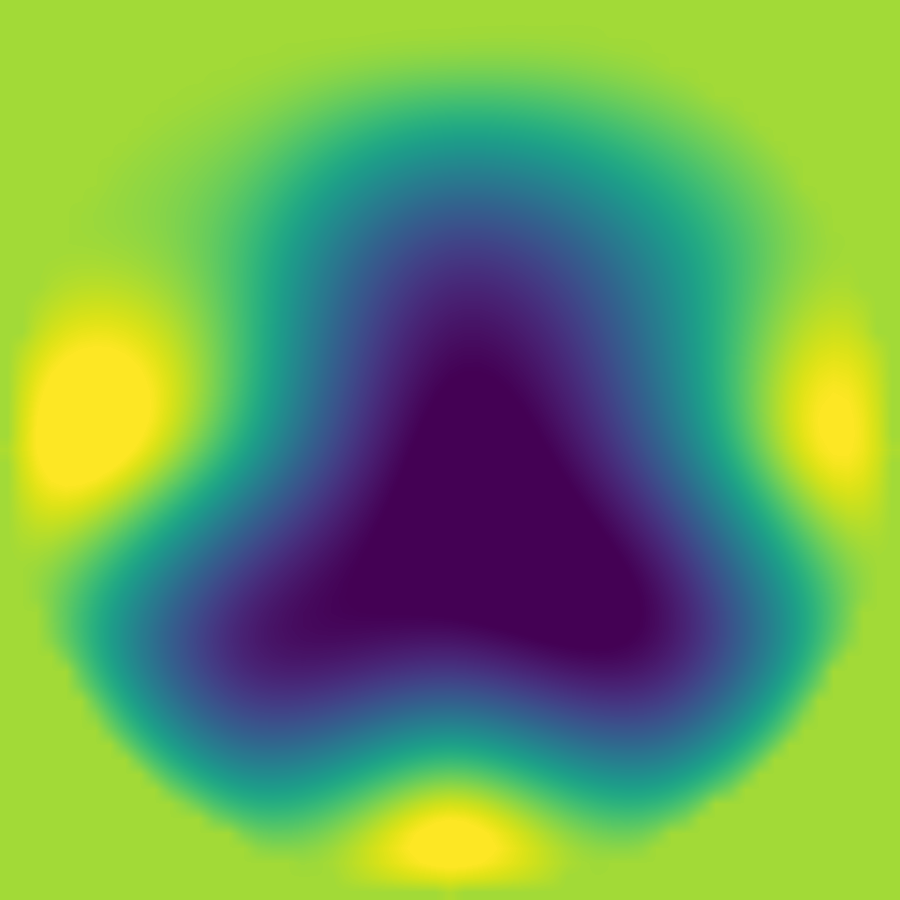} &
  \includegraphics[width=0.178\textwidth]{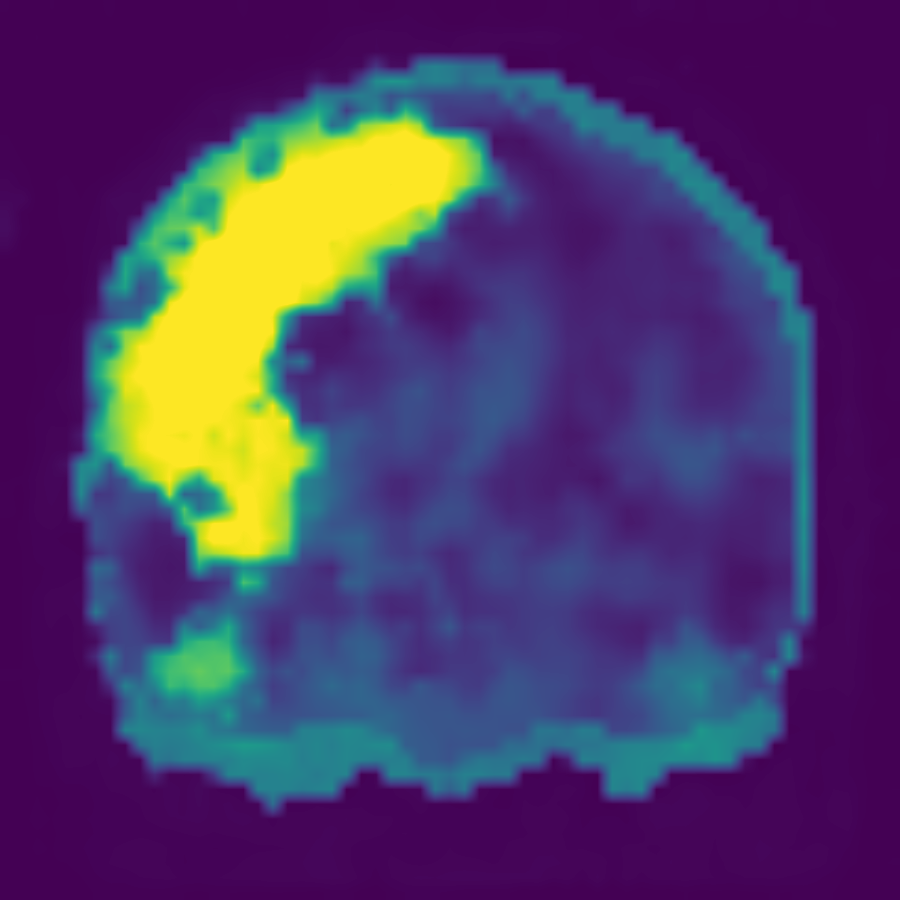} &
  \includegraphics[width=0.178\textwidth]{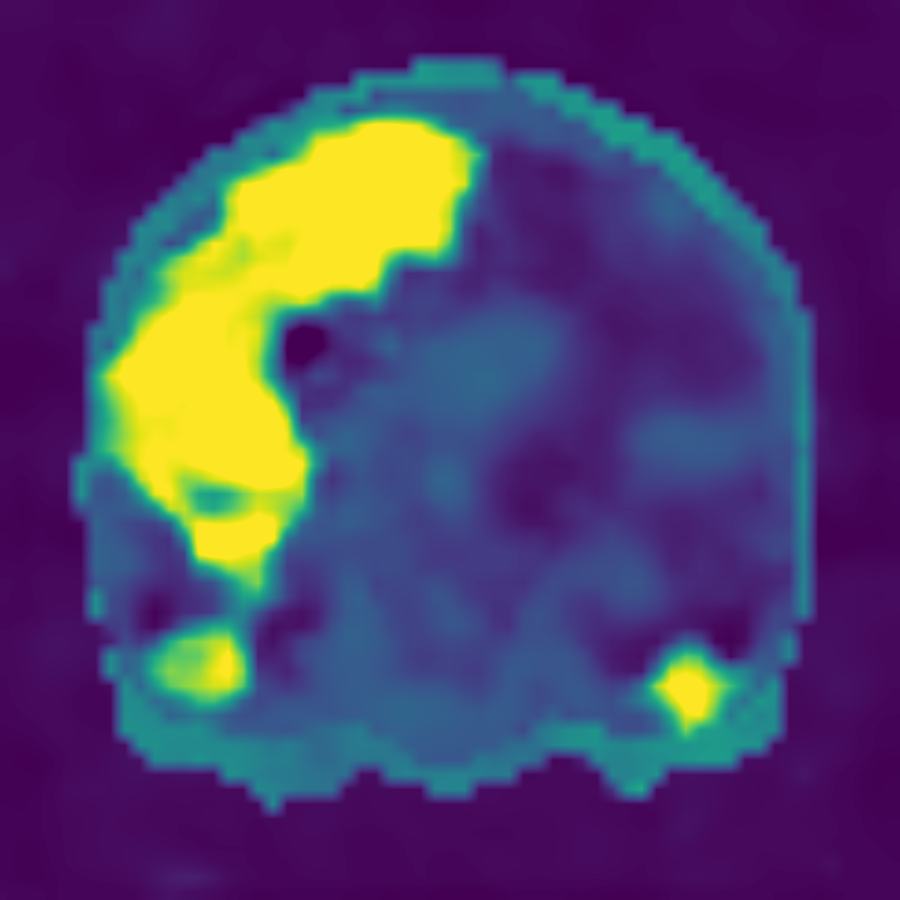} &
  \includegraphics[width=0.178\textwidth]{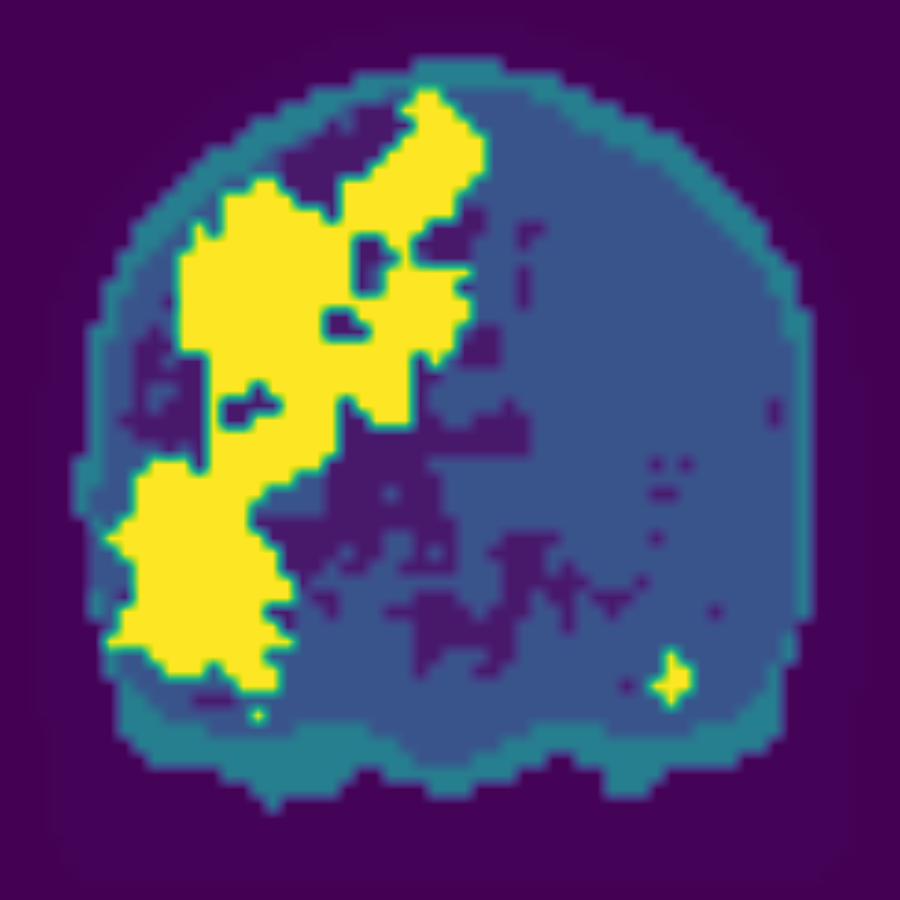} \\
  & 1.80/53.40 & 2.10/20.70 & 44.40/47.90 & \textbf{45.70/47.80} & \\
  & & & & \textcolor{gray}{(Ours)} & \\
\end{tabular}
\caption{\textbf{Visual comparison} on an ischemic reconstruction case at $f=200\text{--}5$ Hz with $50$\,dB noise. SSIM/CC shown below.}
\label{fig:visual}
\end{figure}

\begin{table}[t]
    \centering
    \setlength{\tabcolsep}{4pt}
    \caption{
        \textbf{Quantitative evaluation and efficiency comparison} of dFNOBar vs.
        state-of-the-art methods under different noise levels.
        SSIM and CC $\times 100$. Memory (GB); inference time (s).
        \textbf{Bold}: best; \underline{underlined}: second best.
    }
    \label{tab:comparison_refined}

    {\fontsize{8}{9}\selectfont
    \begin{tabular}{lccccccccc}
        \toprule
        \multicolumn{1}{c}{} &
        \multicolumn{2}{c}{Noiseless} &
        \multicolumn{2}{c}{$50$dB} &
        \multicolumn{2}{c}{$40$dB} &
        \multicolumn{3}{c}{Efficiency} \\
        \cmidrule(r){2-3}
        \cmidrule(r){4-5}
        \cmidrule(r){6-7}
        \cmidrule(r){8-10}

        Method &
        SSIM$^\uparrow$ & CC$^\uparrow$ &
        SSIM$^\uparrow$ & CC$^\uparrow$ &
        SSIM$^\uparrow$ & CC$^\uparrow$ &
        Params & Mem. & Inf. \\
        \midrule

        \multicolumn{10}{l}{\textit{Classical}} \\

        One step GN~\cite{noser}
            & -0.06 & -6.74
            & -0.02 & -4.39
            & -0.02 & -4.28
            & -- & 0.045 & 2.499 \\

        DBar~\cite{nachman1988reconstructions}
            & 0.25 & 15.94
            & 0.21 & 15.20
            & 0.20 & 14.88
            & -- & 0.492 & 0.174 \\

        \midrule

        \multicolumn{10}{l}{\textit{Deep Learning}} \\

        Deep DBar~\cite{deepdbar}
            & \underline{40.09} & \underline{47.66}
            & \underline{37.46} & \textbf{46.64}
            & \underline{29.56} & \textbf{40.6}
            & 90.30M & 0.689 & 0.227 \\

        dFNOBar \textcolor{gray}{(Ours)}
            & \textbf{44.13} & \textbf{50.54}
            & \textbf{39.59} & \underline{46.57}
            & \textbf{32.48} & \underline{40.51}
            & 67.21M & 0.595 & 0.321 \\

        \bottomrule
    \end{tabular}
    }
\end{table}

\section{Experiments}

\subsection{Experimental Setup}
\subsubsection{Dataset and Evaluation metrics.}
The public UCLH release does not include usable voltages on the drive electrodes required by electrode-data 3D D-bar; therefore we evaluate on fully specified synthetic data (ischemic $n{=}250$, hemorrhagic $n{=}219$, healthy $n{=}193$). Simulation uses complex admittivity $\gamma(\omega,x)$, but reconstructions and learning targets predict conductivity $\sigma{=}\Re({\gamma})$. Samples use frequency-difference with reference $f_0{=}5$~Hz paired with all other frequencies; we split patient-wise (90/10) before generating frequency/mode samples, ensuring no patient appears in both sets. Metrics are computed in the brain region (excluding the constant skull, scalp, and empty areas) and we report SSIM and CC across three noise levels.

\begin{figure}[t]
    \centering
    \includegraphics[width=0.49\textwidth]{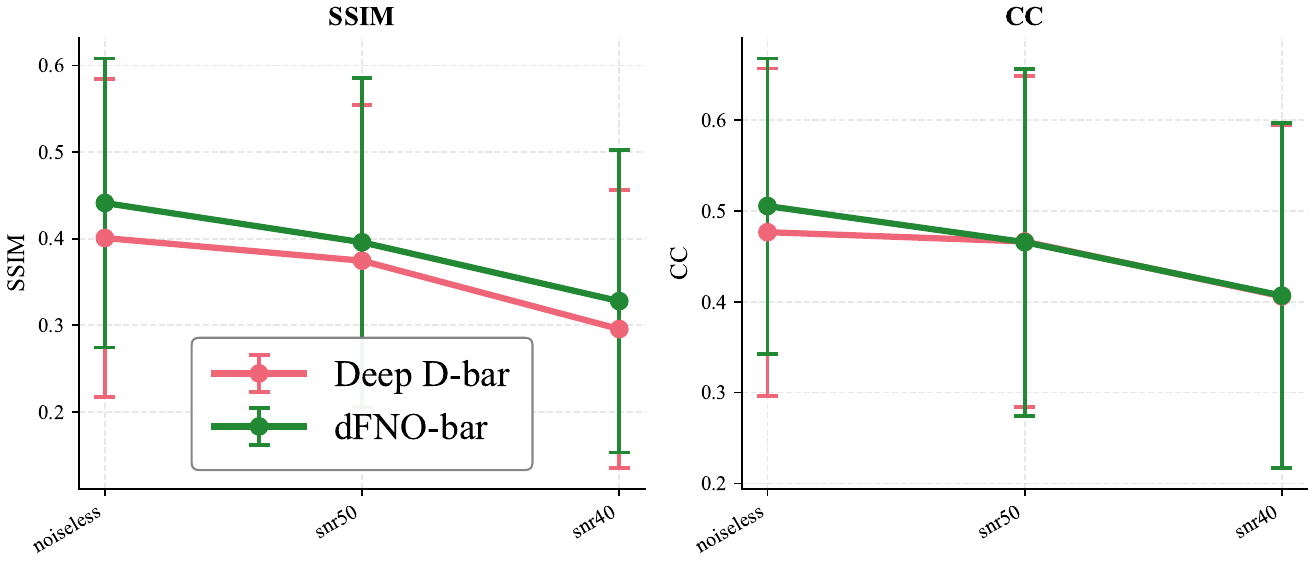}
    \includegraphics[width=0.49\textwidth]{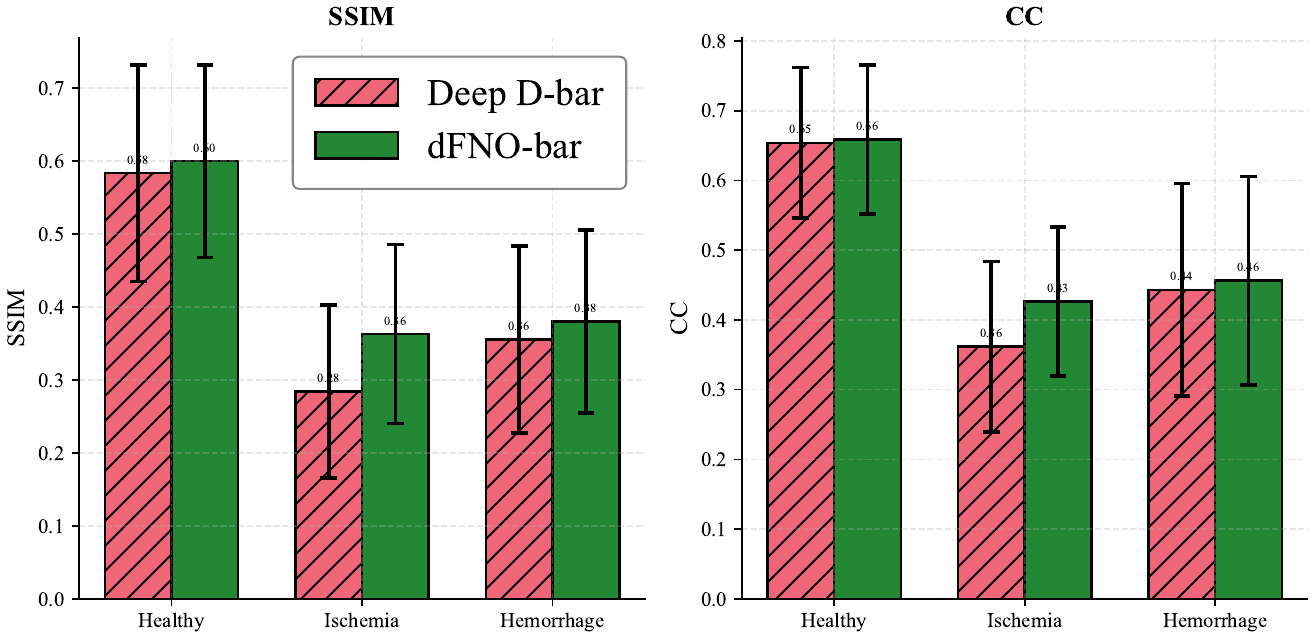}
    \caption{\textbf{Quantitative evaluation of reconstruction performance}: left across noise levels; right across lesion types.}
    \label{fig:fig15}
\end{figure}

\subsubsection{State-of-the-art baselines.}
We compare against classical baselines: (i) \textbf{D-bar}~\cite{nachman1988reconstructions} with $t_0$ at $T_\xi{=}7$, and (ii) \textbf{one-step GN}~\cite{noser}. As learning-based baseline we use \textbf{Deep D-bar}~\cite{deepdbar}, a U-Net post-processor on top of D-bar.

\subsubsection{Implementation and training details.}
We train two learned reconstruction models for 50 epochs and batch size of 14. Deep D-bar~\cite{deepdbar} uses the $t_0$-based D-bar reconstruction at $T_\xi{=}7$ as input and learns a U-Net postprocessor. \textbf{dFNO-bar} takes the complex scattering $t_0$ directly at four truncations $T_\xi{\in}\{5,7,9,11\}$, stacks real/imaginary parts into an 8-channel input, and predicts conductivity using a 6-layer 3D FNO (32 ch. $n\_modes{=}32^3$) followed by a lightweight U-Net refinement head (32, 64, 128, 256). We also compare with the one step Gauss--Newton~\cite{noser}, by calculating the Jacobian at a homogeneous $\sigma_0{=}0.2$ as a rough adaptation to do the frequency-difference.

\subsection{Comparison with state-of-the-art methods}
Table~\ref{tab:comparison_refined} and Fig.~\ref{fig:fig15} report masked SSIM/CC across noise levels and lesion types.
Paired tests (dFNO-bar vs Deep D-bar; n=1088) show significant SSIM gains  for all noise levels ($p{\ll}0.01$; mean $+3.1\%$ across noise settings), whereas CC improvements are significant only in the noiseless setting and not at 40/50\,dB.

\subsubsection{Quantitative and Visual comparisons.}
dFNO-bar improves SSIM consistently over Deep D-bar and strongly outperforms classical D-bar and one-step GN. Qualitative comparisons in Fig.~\ref{fig:visual} demonstrate that dFNO-bar yields more accurate predictions of ischemic lesion location, lesion shape, and brain structure than the competing methods.

\subsubsection{Efficiency comparison.}
We report parameter count, peak memory, and per-sample inference time in Table~\ref{tab:comparison_refined}, both models were trained at 16-mixed precision. dFNO-bar achieves higher accuracy than Deep D-bar with fewer parameters and lower memory, at slightly higher runtime.

\section{Conclusion}

We presented \textit{BREIT}, a modular and reproducible pipeline for 3D MF-EIT stroke reconstruction, spanning neuroimaging-to-admittivity ground-truth generation, a Python-based 3D CEM forward solver, and a 3D D-bar implementation that supports non-uniform electrode layouts. Building on this, we introduced \textit{dFNO-bar} which learns a scattering-to-conductivity mapping by combining multi-truncation D-bar scattering data with a 3D Fourier Neural Operator and a lightweight U-Net refinement head. On our synthetic dataset generated with BREIT, dFNO-bar achieved consistently higher brain SSIM and comparable CC to Deep D-bar across noise settings, while using fewer parameters and memory. Limitations and future work include clinical evaluation via partial-boundary methods to handle missing drive-electrode voltages, and extending data generation to a wider range of subject-specific anatomies. We expect BREIT to facilitate standardized comparisons and accelerate development of robust 3D MF-EIT reconstruction methods for time-sensitive stroke assessment.

\bibliographystyle{splncs04}
\bibliography{biblio}

\end{document}